\documentclass[lettersize,journal,twoside]{IEEEtran}
\usepackage{amsmath,amsfonts,amssymb}
\usepackage{algorithm}
\usepackage{algpseudocode}
\usepackage{array}
\usepackage[caption=false,font=normalsize,labelfont=sf,textfont=sf]{subfig}
\usepackage{textcomp}
\usepackage{stfloats}
\usepackage{url}
\usepackage{verbatim}
\usepackage{graphicx}
\usepackage{cite}
\usepackage[export]{adjustbox}
\usepackage[table,xcdraw]{xcolor}
\usepackage{color}
\usepackage[hidelinks]{hyperref}
\definecolor{linkcolor}{HTML}{ee0000}
\hypersetup{
     colorlinks=true,
	 linkcolor=linkcolor,
	 filecolor=linkcolor,
	 urlcolor=linkcolor,
	 citecolor=linkcolor
}
\usepackage{multirow}
\usepackage{booktabs}
\usepackage{orcidlink}
\usepackage{threeparttable}
\usepackage{pifont}

\newcommand\mlcomment[1]{\iffalse #1 \fi}
\newcommand\bsm[1]{\boldsymbol{\mathrm{#1}}}
\newcommand\transform[2]{{\bsm{T}_{#1}^{#2}}}
\newcommand\transformhat[2]{{\hat{\bsm{T}}_{#1}^{#2}}}
\newcommand\transformtilde[2]{{\tilde{\bsm{T}}_{#1}^{#2}}}
\newcommand\rotation[2]{{\bsm{R}_{#1}^{#2}}}
\newcommand\rotationhat[2]{{\hat{\bsm{R}}_{#1}^{#2}}}

\newcommand\timeoffset[2]{{\tau_{#1}^{#2}}}
\newcommand\timeoffsethat[2]{{\hat{\tau}_{#1}^{#2}}}

\newcommand\translation[2]{{\bsm{p}_{#1}^{#2}}}
\newcommand\translationhat[2]{{\hat{\bsm{p}}_{#1}^{#2}}}

\newcommand\smallminus{{\text{-}}}
\newcommand\smallplus{{\text{+}}}
\newcommand\coordframe[1]{\underrightarrow{\mathcal{F}}_{#1}}
\newcommand{\tabtitlespace}{\vspace{-5pt}}

\newcommand{\figtabbottomspace}{\vspace{-15pt}}
\usepackage{siunitx}
\begin{document}

\title{eKalibr-Stereo: Continuous-Time Spatiotemporal Calibration for  Event-Based Stereo Visual Systems}
\author{
Shuolong Chen \hspace{-1mm}$^{\orcidlink{0000-0002-5283-9057}}$, Xingxing Li \hspace{-1mm}$^{\orcidlink{0000-0002-6351-9702}}$, and Liu Yuan \hspace{-1mm}$^{\orcidlink{0009-0003-6039-7070}}$

\thanks{
This work was supported by the National Science Fund for Distinguished Young Scholars of China under Grant 42425401.}
\thanks{The authors are with the School of Geodesy and Geomatics (SGG), Wuhan University (WHU), Wuhan 430070, China.
Corresponding author: Xingxing Li (\texttt{xxli@sgg.whu.edu.cn}). 
The specific contributions of the authors to this work are listed in Section \hyperref[sect:author_contribution]{\textbf{CRediT Authorship Contribution Statement}} at the end of the article.}	
}


\markboth{Journal of \LaTeX\ Class Files,~Vol.~14, No.~8, August~2021}
{Chen \MakeLowercase{\textit{et al.}}: eKalibr-Stereo: Continuous-Time Spatiotemporal Calibration for  Event-Based Stereo Visual Systems}


\maketitle

\begin{abstract}
The bioinspired event camera, distinguished by its exceptional temporal resolution, high dynamic range, and low power consumption, has been extensively studied in recent years for motion estimation, robotic perception, and object detection.
In ego-motion estimation, the stereo event camera setup is commonly adopted due to its direct scale perception and depth recovery.
For optimal stereo visual fusion, accurate spatiotemporal (extrinsic and temporal) calibration is required.
In this letter, we present \emph{eKalibr-Stereo}, an accurate spatiotemporal calibrator for event-based stereo visual systems, utilizing the widely used circular grid board.
To ensure the continuity of grid pattern tracking, building upon the grid pattern recognition method in \emph{eKalibr}, an additional motion prior-based tracking module is designed in \emph{eKalibr-Stereo} to track incomplete grid patterns.
Based on tracked grid patterns, a two-step initialization procedure is performed to recover initial guesses of piece-wise B-splines and spatiotemporal parameters, followed by a continuous-time batch bundle adjustment to refine the initialized states to optimal ones.
The results of extensive real-world experiments show that \emph{eKalibr-Stereo} can achieve accurate event-based stereo spatiotemporal calibration.
The implementation of \emph{eKalibr-Stereo} is open-sourced at (\url{https://github.com/Unsigned-Long/eKalibr}) to benefit the research community.
\end{abstract}


\section{Introduction and Related Works}
\IEEEPARstart{B}{ioinspired} event cameras have attracted considerable research interest in recent years, due to their advantages of low sensing latency and high dynamic range over conventional standard (frame-based) cameras \cite{guan2023pl}.
The ego-motion estimation in high-dynamic-range and high-speed scenarios is one of applications of the event camera, where a stereo camera setup is commonly employed for direct scale recovery \cite{chen2023esvio,zhou2021event,LU-RSS-24}.
For such an event-based stereo visual sensor suite, accurate spatiotemporal calibration is required to determine extrinsics and time offset between cameras for subsequent data fusion.

\begin{figure}[t]
\centering
\includegraphics[width=\linewidth]{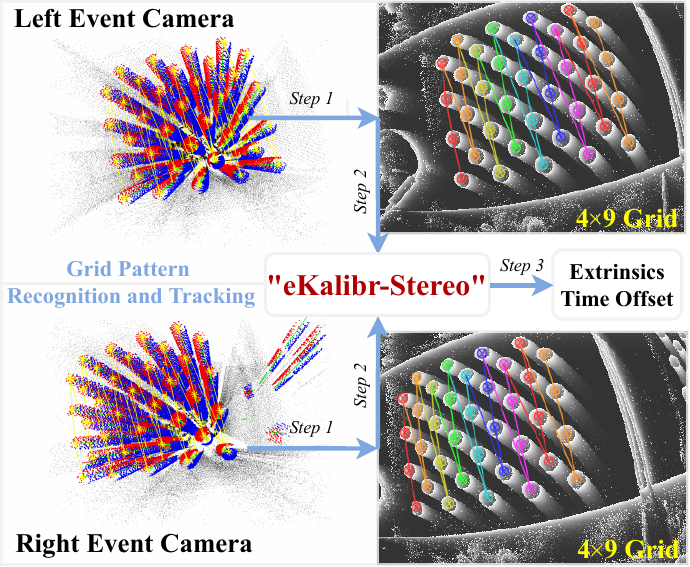}
\caption{The runtime visualization of \emph{eKalibr-Stereo}.
\emph{eKalibr-Stereo} tracks grid patterns using raw events of stereo event cameras and estimates spatiotemporal (extrinsic and temporal) parameters using continuous-time-based batch optimization.}
\label{fig:ekalibr}
\figtabbottomspace
\end{figure}

Stereo visual spatiotemporal calibration typically consists of two sub-modules: ($i$) correspondence construction (front end) and ($ii$) spatiotemporal optimization (back end).
In the front end, artificial visual targets, such as checkerboards \cite{yu2006robust}, April Tags \cite{wang2016apriltag},  and ChArUco board \cite{an2018charuco}, are commonly employed to construct accurate 3D-2D correspondences with real-world geometric scale through pattern recognition.
While a substantial number of target pattern recognition methods \cite{yu2006robust,sun2008robust,hu2019deep} oriented to standard cameras have been proposed, they are not applicable to event cameras, which output
asynchronous event stream rather than conventional intensity images.
To recognize target patterns from raw events, early works \cite{rpg_dvs_ros_calibration,dominguez2019bio,cai2024accurate} generally rely on blinking light emitting diode (LED) grid boards.
Although target patterns can be accurately extracted, requiring additional LED boards introduces inconvenience.
Meanwhile, these methods typically require the event camera to remain stationary, making them unsuitable for multi-camera spatiotemporal calibration that necessitates motion excitation \cite{chen2025ikalibr}.
To address this, subsequent methods \cite{muglikar2021calibrate,jiao2023lce} have proposed an alternative approach, namely reconstructing intensity images from raw events using event-based image reconstruction methods (such as E2VID\cite{rebecq2019high} and Spade-E2VID \cite{cadena2021spade}) first, followed by conventional image-based pattern recognition methods.
Although reconstructed images exhibit high consistency, substantial noise within the images could lead to imprecise pattern extraction, which further affects calibration accuracy.
Considering these, some event-based pattern recognition methods have been proposed recently, aiming to extract target patterns from dynamically acquired raw events directly.
Based on density-based spatial clustering (DBSCAN), Huang et al. \cite{huang2021dynamic} cluster events accumulated in a time window and fit circles to extract centers of circles on a circle grid board.
Similarly but performing DBSCAN in spatiotemporal domain, Salah et al. \cite{salah2024calib} cluster events generated from a circle grid board, and then fit 3D cylinders for circle center determination.
DBSCAN-based clustering methods are sensitive to hyperparameters, which can lead to instability in grid pattern recognition and further affect the calibration.
Differently and more directly, the authors of \emph{EF-Calib} \cite{wang2024ef} specialized in target design and developed a novel circular grid board with cross points, which enables efficient and accurate event-based (using circle edges) and frame-based (using cross points) target recognition for subsequent event-frame stereo camera calibration.
In this method, events of circle edges are clustered using BBDT and then matched based on cluster distance.
Similarly, the authors of the event-based intrinsic calibrator \emph{eKalibr} \cite{chen2025ekalibr} proposed an event-only target pattern recognition method designed for the commonly used circular grid board.
This method clusters and matches event clusters based on normal flow estimation (from first principles of events), rather than relying on traditional algorithms like BBDT in \cite{wang2024ef}, resulting in more efficient and explainable extraction.

In terms of the back end of the stereo visual calibration, namely spatiotemporal optimization, event-based and frame-based calibrations share the same algorithmic framework, aiming to estimate spatiotemporal parameters using extracted visual target patterns.
In general, spatiotemporal optimization can be categorized into discrete-time-based and continuous-time-based ones.
Discrete-time-based methods represent states using discrete estimates that are temporally coupled to measurements.
Based on the extended Kalman filter (EKF), Mirzaei et al. \cite{mirzaei2008kalman} proposed a visual-inertial extrinsic calibration method to determine the transformation between a standard camera and an inertial measurement unit (IMU).
Similarly, Hartzer et al. \cite{hartzer2022online} presented an EKF-based online visual-inertial extrinsic calibration method.
Yang et al. \cite{yang2016monocular} designed a sliding-window-based visual-inertial state estimator, supporting online camera-IMU extrinsic calibration.
Different from the discrete-time-based methods, continuous-time-based ones represent time-varying states using time-continuous functions (such as B-splines), enabling state querying at any time instance, and thus are more suitable for temporal calibration.
The well-known \emph{Kalibr} \cite{furgale2013unified} proposed by Furgale et al. is the first continuous-time-based calibration framework, which employs B-splines for state representation and supports both extrinsic and temporal calibration for visual-inertial, multi-IMU, and multi-camera sensor suites.
\emph{Kalibr} is then extended by Huai et al. \cite{huai2022continuous} to support the rolling shutter cameras for readout time calibration.
In addition to vision-related calibration, the continuous-time state representation has also been widely employed in other multi-sensor calibration, such as LiDAR-IMU \cite{lv2022observability} and radar-IMU \cite{chen2024ris} calibration.

In this article, focusing on event-based stereo visual systems, we present a continuous-time-based spatiotemporal calibration method, named \emph{eKalibr-Stereo}, to accurately estimate the extrinsics and time offset between event cameras.
Building upon \emph{eKalibr} \cite{chen2025ekalibr}, \emph{eKalibr-Stereo} tracks continuous circle grid patterns (complete and incomplete ones) from raw events for 3D-2D correspondence construction.
Given the high non-linearity of continuous-time optimization, a two-stage initialization procedure is first conducted to recover the initials of states, which are then iteratively refined to optimal ones using a continuous-time-based batch bundle adjustment.
\emph{eKalibr-Stereo} makes the following (potential) contributions:
\begin{enumerate}
\item We proposed a continuous-time-based spatial and temporal calibrator for event-based stereo visual systems, which could accurately determine both extrinsics and time offset of a stereo event camera system.
To the best of our knowledge, this is the first open-source work focused on event-based stereo spatiotemporal calibration.

\item We designed a motion-prior-aided tracking module for incomplete grid pattern identification, to maximize the continuity of pattern tracking, facilitating the final spatiotemporal optimization.

\item Sufficient real-world experiments were conducted to comprehensively evaluate the proposed \emph{eKalibr-Stereo}.
Both the dataset and code implementation are open-sourced, to benefit the robotic community if possible.
\end{enumerate}

Note that the proposed \emph{eKalibr-Stereo} supports \textbf{one-shot} event-based \textbf{multi-camera} spatiotemporal calibration (an arbitrary number of event cameras).
To enhance clarity, this article only considers the minimal stereo event camera configuration, as it's the most typical sensor setup for facilitating multi-camera calibration.

\section{Preliminaries}
This section presents notations and definitions utilized in this article.
The involved camera intrinsic model and B-spline-based time-varying state representation are also introduced for a self-contained exposition of this work.

\subsection{Notations and Definitions}
Given a raw event $\bsm{e}$ generated by the event camera, we use $\tau\in\mathbb{R}$, $\bsm{x}\in\mathbb{Z}^2$, and $p\in\{\smallminus 1, \smallplus 1\}$ to represent its timestamp, pixel position, and polarity, respectively, i.e., $\bsm{e}\triangleq\{\tau,\bsm{x},p\}$.
The camera frame and world frame (defined by the circle grid board) are represented as $\coordframe{c}$ and $\coordframe{w}$, respectively.
The transformation from $\coordframe{c}$ to $\coordframe{w}$ are parameterized as the Euclidean matrix $\transform{c}{w}\in\mathrm{SE(3)}$, which is defined as:
\begin{equation}
\transform{c}{w}\triangleq\begin{bmatrix}
\rotation{c}{w}&\translation{c}{w}\\
\bsm{0}_{1\times 3}&1
\end{bmatrix}
\end{equation}
where $\rotation{c}{w}\in\mathrm{SO(3)}$ and $\translation{c}{w}\in\mathbb{R}^3$ are the rotation matrix and translation vector, respectively.
Finally, we use $\hat{(\cdot)}$ and $\tilde{(\cdot)}$ to represent the state estimates and noisy quantities (e.g., the generated raw events and extracted grid patterns), respectively.

\subsection{Camera Intrinsic Model}
The camera intrinsic model characterizes the visual projection process whereby 3D points in the camera coordinate frame are geometrically mapped onto the image plane to derive corresponding 2D pixels.
Adhering to our previously proposed \emph{eKalibr} \cite{chen2025ekalibr},  the intrinsic camera model comprising the pinhole projection model \cite{kannala2006generic} and radial-tangential distortion model \cite{tang2017precision} are employed in this work, which can be expressed as:
\begin{equation}
\label{equ:proj_func}
\bsm{x}_p=\pi\left(\bsm{p}^c,\mathcal{X}_{\mathrm{intr}}\right)\triangleq
\bsm{K}\left(\mathcal{X}_{\mathrm{proj}}\right)\cdot 
\bsm{d}\left(\bsm{p}^c,\mathcal{X}_{\mathrm{dist}}\right)
\end{equation}
with
\begin{equation}
\begin{gathered}
\mathcal{X}_{\mathrm{intr}}\triangleq\mathcal{X}_{\mathrm{proj}}\cup\mathcal{X}_{\mathrm{dist}}\\
\mathcal{X}_{\mathrm{proj}}\triangleq\left\lbrace
f_x,f_y,c_x,c_y
\right\rbrace,\;
\mathcal{X}_{\mathrm{dist}}\triangleq\left\lbrace
k_1,k_2,p_1,p_2
\right\rbrace
\end{gathered}
\end{equation}
where $\bsm{d}:\mathbb{R}^{3}\mapsto\mathbb{R}^{3}$ represents the distortion function distorting normalized image coordinates using distortion parameters $\mathcal{X}_{\mathrm{dist}}$;
$\bsm{K}\in\mathbb{R}^{2\times 3}$ denotes the intrinsic matrix organized by projection parameters $\mathcal{X}_{\mathrm{proj}}$;
$\pi:\mathbb{R}^{3}\mapsto\mathbb{R}^2$ is the projection function projecting 3D point $\bsm{p}^c$ onto the image plane as 2D point $\bsm{x}_{p}$;
$\mathcal{X}_{\mathrm{intr}}$ represents the intrinsic parameters comprising $\mathcal{X}_{\mathrm{proj}}$ and $\mathcal{X}_{\mathrm{dist}}$, which can be pre-calibrated using \emph{eKalibr}.

\subsection{Continuous-Time State Representation}
To efficiently fuse asynchronous data for multi-sensor spatiotemporal determination, especially for time offset calibration, the continuous-time state representation is employed in this work to represent the time-varying rotation and position of event cameras.
Compared with the conventional discrete-time representation generally maintaining discrete states at measurement times, the continuous-time representation models time-varying states using time-continuous functions, such as Gaussian process regression \cite{barfoot2014batch}, hierarchical wavelets \cite{anderson2014hierarchical}, and B-splines \cite{furgale2012continuous}, enabling state querying at arbitrary time.
In this work, the uniform B-spline is utilized for continuous-time state representation, which inherently possesses sparsity due to its local controllability, allowing computation acceleration in optimization \cite{chen2025ikalibr}.

The uniform B-spline is characterized by the spline order, a temporally uniformly distributed control point sequence, and a constant time distance between neighbor control points.
Specifically, given a series of translational control points:
\begin{equation}
\label{equ:pos_cp}
\begin{gathered}
\mathcal{X}_{\mathrm{pos}}\triangleq\left\lbrace
\bsm{p}_i,\tau_i\mid\bsm{p}_i\in\mathbb{R}^3,\tau_i\in\mathbb{R}
\right\rbrace 
\\
\mathrm{s.t.}\;\;
\tau_{i\smallplus 1}-\tau_i\equiv\Delta\tau_{\mathrm{pos}}
\end{gathered}
\end{equation}
the position $\bsm{p}(\tau)$ at time $\tau\in[\tau_i, \tau_{i\smallplus 1})$ of a $k$-order uniform B-spline can be computed as follows:
\begin{equation}
\label{equ:pos_bspline}
\begin{gathered}
\bsm{p}(\tau)=\bsm{p}_i+\sum_{j=1}^{k\smallplus 1}\lambda_j(u)\cdot\left(\bsm{p}_{i\smallplus j}-\bsm{p}_{i\smallplus j\smallminus 1} \right) 
\\
\mathrm{s.t.}\;\;
u=\frac{\tau-\tau_i}{\Delta\tau_{\mathrm{pos}}}
\end{gathered}
\end{equation}
where $\lambda_j(\cdot)$ denotes the $j$-th element of vector $\bsm{\lambda}(u)$ obtained from the order-determined cumulative
matrix and $u$ \cite{chen2025ikalibr}.
In this work, the cubic uniform B-spline ($k=4$) is employed.

The B-spline representation of time-varying rotation has similar forms with (\ref{equ:pos_bspline}) by replacing vector addition in $\mathbb{R}^3$ with group multiplication in $\mathrm{SO(3)}$.
The key distinction resides in the scalar multiplication operated within the Lie algebra $\mathfrak{so}(3)$, rather than on the Lie group manifold, to ensure closedness \cite{sommer2020efficient}.
Specifically, given a series of rotational control points:
\begin{equation}
\label{equ:rot_cp}
\begin{gathered}
\mathcal{X}_{\mathrm{rot}}\triangleq\left\lbrace
\bsm{R}_i,\tau_i\mid\bsm{R}_i\in\mathrm{SO(3)},\tau_i\in\mathbb{R}
\right\rbrace 
\\
\mathrm{s.t.}\;
\tau_{i\smallplus 1}-\tau_i\equiv\Delta\tau_{\mathrm{rot}}
\end{gathered}
\end{equation}
the rotation $\bsm{R}(\tau)$ at time $\tau\in[\tau_i, \tau_{i\smallplus 1})$ of a $k$-order uniform B-spline can be computed as follows:
\begin{equation}
\label{equ:rot_bspline}
\begin{gathered}
\bsm{R}(\tau)=\bsm{R}_i\cdot\prod_{j=1}^{k\smallplus 1}\mathrm{Exp}\left( \lambda_j(u)\cdot\mathrm{Log}\left(\bsm{R}_{i\smallplus j\smallminus 1}^\top\cdot\bsm{R}_{i\smallplus j} \right)\right)  
\\
\mathrm{s.t.}\;\;
u=\frac{\tau-\tau_i}{\Delta\tau_{\mathrm{rot}}}
\end{gathered}
\end{equation}
where $\mathrm{Exp}(\cdot)$ maps elements in the Lie algebra to the associated Lie group, and $\mathrm{Log}(\cdot)$ is its inverse operation.

\section{Methodology}
This section presents the proposed event-based continuous-time stereo visual spatiotemporal calibration framework.

\subsection{System Overview}
\label{sect:overview}
\begin{figure}[t]
	\centering
	\includegraphics[width=\linewidth]{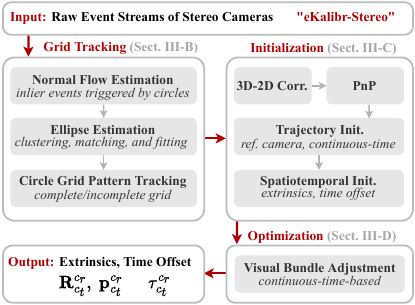}
	\caption{Illustration of the pipeline of the proposed event-based stereo visual spatiotemporal calibration method. A detailed description of the pipeline is provided in Section \ref{sect:overview}.}
	\label{fig:overview}
	\figtabbottomspace
\end{figure}

The comprehensive framework of the proposed event-based stereo visual calibrator is illustrated in Fig. \ref{fig:overview}.
Given raw asynchronous event streams from the stereo camera rig, we first perform normal flow estimation and ellipse fitting for each event camera, to track both complete and incomplete circle grid patterns, see Section \ref{sect:grid_tracking}.
The event camera that tracks the grid pattern the most among two cameras would be treated as the reference (primary) camera (denoted as $\coordframe{c_r}$), while the other would be assigned as the target (secondary) camera (denoted as $\coordframe{c_t}$).
Subsequently, for each tracked grid pattern of the reference camera, PnP \cite{lepetit2009ep} would be employed to estimate the camera pose based on 3D-2D correspondences.
Discrete poses of the reference camera are then utilized to recover a continuous-time trajectory, see Section \ref{sect:traj_init}.
The spatiotemporal parameters, i.e., extrinsics and time offset, would be also initialized based on continuous-time hand-eye alignment, see Section \ref{sect:hand_eye}.
Finally, a continuous-time-based visual bundle adjustment would be performed to refine all states in the estimator to the global optimal ones, see Section \ref{sect:batch_opt}.

The state vector of the system can be described as follows:
\begin{equation}
\mathcal{X}\triangleq\left\lbrace\mathcal{X}_{\mathrm{pos}},\mathcal{X}_{\mathrm{rot}},\rotation{c_t}{c_r},\translation{c_t}{c_r},\timeoffset{c_t}{c_r}\right\rbrace
\end{equation}
where $\mathcal{X}_{\mathrm{pos}}$ and $\mathcal{X}_{\mathrm{rot}}$ are translational and rotational control points defined in (\ref{equ:pos_cp}) and (\ref{equ:rot_cp}), respectively;
$\rotation{c_t}{c_r}$ and $\translation{c_t}{c_r}$ denote the extrinsic rotation and translation from $\coordframe{c_t}$ to $\coordframe{c_r}$;
$\timeoffset{c_t}{c_r}$ represents the time offset between two cameras, i.e., temporal transformation $\tau_{c_r}=\tau_{c_t}+\timeoffset{c_t}{c_r}$ holds.
The extrinsics and time offset are exactly the spatiotemporal parameters \emph{eKalibr-Stereo} calibrates.

\subsection{Event-Based Circle Grid Tracking}
\label{sect:grid_tracking}
Given generated raw event streams, we first employ the event-based circle grid pattern recognition algorithm \cite{chen2025ekalibr} proposed in  \emph{eKalibr} to extract \textbf{complete} grid patterns for each camera.
As described in \cite{chen2025ekalibr}, we first perform event-based normal flow estimation \cite{lu2024EventBased} on the surface of active event (SAE) \cite{delbruck2008frame} and homopolarly cluster inlier events for cluster matching.
Spatiotemporal ellipses would then be estimated for each matched cluster pair for center determination of the grid circle.
Finally, temporally synchronized centers would be organized as ordered grid patterns.
Note that although both the asymmetric and symmetric circle grids are supported in \emph{eKalibr}, the asymmetric circle grid is utilized in this work, as it does not exhibit 180-degree ambiguity \cite{mathworks_calibration_patterns}.

\begin{figure}[t]
	\centering
	\includegraphics[width=\linewidth]{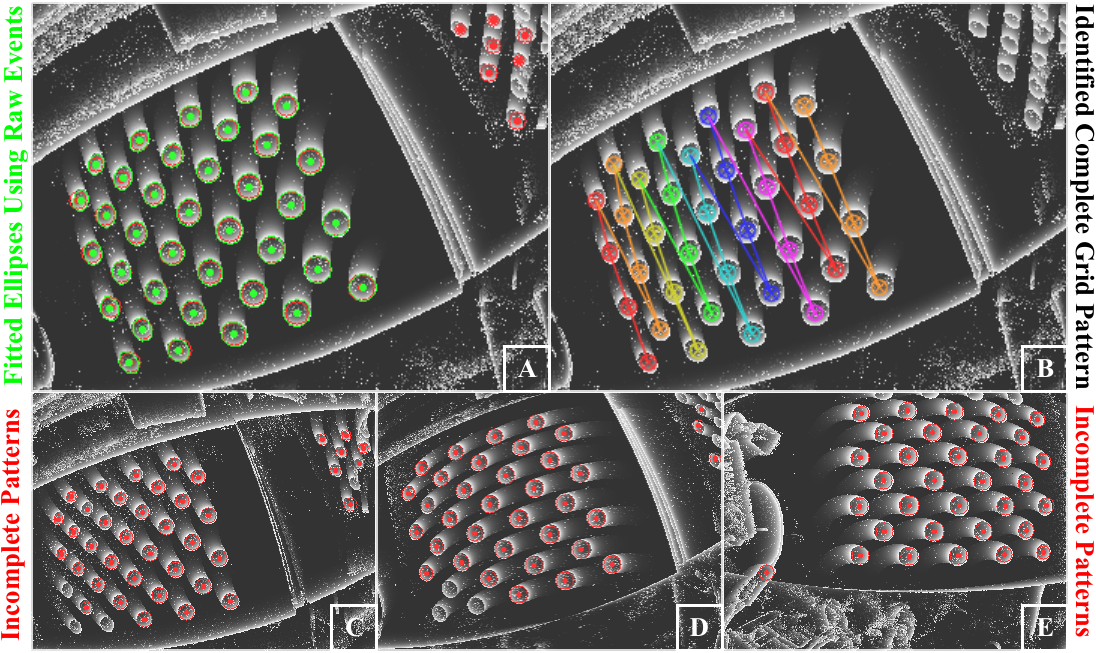}
	\caption{Schematic of incomplete grid pattern tracking.
	When ellipses corresponding to all grid circles are fitted (subfigure \textbf{A}), a complete pattern (subfigure \textbf{B}) can be extracted. However, due to oblique perspective (subfigure \textbf{C}), noisy events (subfigure \textbf{D}), or the grid being outside the sensing range (subfigure \textbf{E}), not all ellipses can be identified, resulting in incomplete pattern.}
	\label{fig:incomp_tracking}
\end{figure}

\begin{algorithm}[t]
\caption{Incomplete Grid Pattern Tracking}
\label{alg:tracking}
\begin{algorithmic}[1]
\State \textbf{Input:} Extracted complete grid patterns $\mathcal{P_{\mathrm{cmp}}}$ and unorganized fitted ellipses $\mathcal{E}$ for incomplete pattern tracking.
\State \textbf{Output:} Temporally-ordered patterns $\mathcal{P}$ including both complete grids $\mathcal{P_{\mathrm{cmp}}}$ and incomplete ones $\mathcal{P_{\mathrm{incmp}}}$.
\State Initialize grid pattern set $\mathcal{P}\gets\mathcal{P_{\mathrm{cmp}}}$, traverse order $i=2$.
\Repeat
\For{$\left(\mathcal{G}^{k\smallminus1},\tau^{k\smallminus1}\right),\left(\mathcal{G}^k,\tau^k\right),\left(\mathcal{G}^{k\smallplus1},\tau^{k\smallplus1}\right)\in\mathcal{P}$}
\For{center $\bsm{p}_j$ tracked three times $\bsm{p}_j^{k\smallminus1},\bsm{p}_j^{k},\bsm{p}_j^{k\smallplus1}$}
\State Predict $\hat{\bsm{p}}_j^{k\smallplus i}$ at time $\tau^{k\smallplus i}$ using (\ref{equ:lagrange_poly}), associate
\Statex\hspace{1.57cm}its nearest ellipse center $\hat{\bsm{p}}_j^{k\smallplus i}\simeq\bsm{c}_n^{k\smallplus i}\in\mathcal{E}^{k\smallplus i}$
\Statex\hspace{1.57cm}if $\Vert\hat{\bsm{p}}_j^{k\smallplus i}-\bsm{c}_n^{k\smallplus i}\Vert_2\le d_{\mathrm{thd}}$, then $\mathcal{G}^{k\smallplus i}\gets\bsm{c}_{j}^{k\smallplus i}$.
\EndFor
\State Store $\mathcal{P}\gets\mathcal{G}^{k\smallplus i}$ if tracked enough points in $\mathcal{G}^{k\smallplus i}$.
\EndFor
\State Reverse order $i\gets \left( -1\right) \times i$.
\Until{no additional incomplete pattern tracked in last loop.}
\State\textbf{Note:} The three-point Lagrange polynomial is defined as:
\begin{equation}
\label{equ:lagrange_poly}
\small
\bsm{p}(\tau)\gets L_3(\tau)\triangleq\sum_{k=0}^{2}\left( \bsm{p}^k\cdot\left( \prod_{l=0,l\ne k}^{2}\frac{\tau-\tau^l}{\tau^k-\tau^l}\right) \right).
\end{equation}
\end{algorithmic}
\end{algorithm}

In addition to the aforementioned grid pattern recognition algorithm \cite{chen2025ekalibr}, this work incorporates an additional module specifically designed to track \textbf{incomplete} grid patterns (see Fig. \ref{fig:incomp_tracking}), to improve continuity of grid tracking\footnote{
\textbf{Continuity of grid tracking}: the motion-based spatiotemporal calibration determines parameters based on the rigid-body constraint under continuous motion, thereby requiring motion state estimation from continuous tracking of the grid.
}.
Specifically, leveraging the prior knowledge of motion continuity, we construct a three-point Lagrange polynomial \cite{werner1984polynomial} for each grid circle that had been continuously tracked three times, and then predict its position in the subsequent SAE map.
When the predicted point exhibits sufficient proximity to its nearest ellipse center extracted from the subsequent SAE, we designate the newly extracted ellipse center as the corresponding position of the grid circle in the subsequent SAE.
Once enough predicted grid circles were associated with ellipse centers in the subsequent SAE map, we organized a new incomplete tracked grid pattern.
Note that, to ensure maximal tracking continuity, we would iteratively perform alternating forward and backward tracking of incomplete grid patterns until no additional ones can be tracked.
The detailed process is summarized in Algorithm \ref{alg:tracking}.
For notational convenience, we denote all tracked grid patterns as:
\begin{equation}
\mathcal{P}\triangleq\left\lbrace
\left( \mathcal{G}^k,\tau^k\right)
\right\rbrace
\;\mathrm{s.t.}\;\;
\mathcal{G}^k\triangleq\left\lbrace
\left. \left( \bsm{x}^k_j,\bsm{p}^w_j\right) \right| 
\bsm{x}^k_j
\in\mathbb{R}^2,\bsm{p}^w_j\in\mathbb{R}^3
\right\rbrace
\end{equation}
where $\mathcal{G}^k$ denotes the $k$-th tracked grid pattern at time $\tau^k$, storing tracked 2D ellipse centers $\left\lbrace \bsm{x}^k_j\right\rbrace $ and their associated 3D grid circle centers $\left\lbrace \bsm{p}^w_j\right\rbrace $ on the board.
The event camera that tracks the grid pattern the most among two cameras is treated as the reference camera (denote its tracked pattern set as $\mathcal{P}_\mathrm{ref}$), while the other is assigned as the target camera (denote its pattern set as $\mathcal{P}_\mathrm{tar}$).

\subsection{Two-Stage State Initialization}
Considering the high non-linearity of continuous-time optimization, an efficient two-stage initialization procedure is designed to orderly recover initial guesses of all parameters in the estimator.

\subsubsection{Continuous-Time Trajectory Initialization}
\label{sect:traj_init}
We first perform PnP \cite{lepetit2009ep} of each tracked grid pattern for both cameras to estimate discrete camera pose sequence, i.e., recover $\{\transform{c_r^k}{w}\}$ using $\mathcal{P}_\mathrm{ref}$ and $\{\transform{c_t^k}{w}\}$ using $\mathcal{P}_\mathrm{tar}$.
Subsequently, we segment poses of the reference camera into multiple sections and then construct piece-wise pose B-splines.
Only those neighbor poses with ($i$) time distances smaller than $\Delta\tau_{\mathrm{thd}}$ and that ($ii$) appear continuously more than $\mathcal{N}_{\mathrm{thd}}$ times, would be considered for B-spline trajectory construction.
Such a strategy is designed to ensure high-precision continuous-time trajectory initialization using sufficient discrete poses.
After segmentation, piece-wise pose B-splines could be initialized by solving the following nonlinear least-squares problem:
\begin{equation}
\label{equ:spline_init}
\small
\hat{\mathcal{X}}_{\mathrm{rot}},\hat{\mathcal{X}}_{\mathrm{pos}}\gets
\arg\min\sum_{k=0}^{n}
\left\|
\mathrm{Log}\left( 
\transformhat{c_r}{w}(\tau_r^k)\cdot
\left( \transformtilde{c_r^k}{w}\right) ^{-1}\right) 
\right\|^2
\end{equation}
where $\transformtilde{c_r^k}{w}$ denotes the estimated $k$-th pose of $\coordframe{c_r}$ using PnP, stamped as time $\tau_r^k$ by the clock of the reference camera;
$\transformhat{c_{r}}{w}(\tau)$ represents the pose at time $\tau$, queried from the corresponding continuous-time trajectory using (\ref{equ:pos_bspline}) for $\translationhat{c_r}{w}(\cdot)$ and (\ref{equ:rot_bspline}) for $\rotationhat{c_r}{w}(\cdot)$, which exactly involves the optimization of control points of B-splines.
Note that for notational simplicity, the B-spline index is omitted in (\ref{equ:spline_init}) when performing pose querying.
Readers should be aware that the pose is queried from the B-spline temporally associated with it (i.e., the timestamp of the pose to query falls in the time interval of associated B-spline).

\subsubsection{Spatiotemporal Initialization}
\label{sect:hand_eye}
After the continuous-time trajectories of the reference camera are initialized, we employ the continuous-time hand-eye alignment\footnote{\textbf{Hand-eye alignment} refers to the process of recovering the spatiotemporal parameters between different sensors through aligning their kinematics based on rigid-body constraints.} \cite{chen2025ikalibr} to initialize the extrinsics and time offset of the target camera with respect to the reference camera. This could be achieved by solving the following least-squares problem:
\begin{equation}
\label{equ:st_init}
\small
\rotationhat{c_t}{c_r},\translationhat{c_t}{c_r},\timeoffsethat{c_t}{c_r}
\!\gets\!\arg\min\!\sum_{k=0}^{n}
\left\|
\mathrm{Log}\!\left(
\transformhat{c_t^{k\smallplus 1}}{c_t^{k}}\!\cdot\!
\left(\transformtilde{c_t^{k\smallplus 1}}{w}\!\right)^{-1}\!\cdot\!
\transformtilde{c_t^k}{w}\right) 
\right\|^2
\end{equation}
with
\begin{equation}
\label{equ:rel_pose}
\small
\transformhat{c_t^{k\smallplus 1}}{c_t^{k}}=
\left( \transformhat{c_t}{c_r}\right)^{-1}\!\cdot
\left( \transform{c_r}{w}(\tau_t^k+\timeoffsethat{c_t}{c_r})\right)^{-1}\!\cdot
\transform{c_r}{w}(\tau_t^{k\smallplus 1}+\timeoffsethat{c_t}{c_r})\cdot
\transformhat{c_t}{c_r}
\end{equation}
where $\transformtilde{c_t^k}{w}$ and $\transformtilde{c_t^{k\smallplus 1}}{w}$ are two consecutive poses of the target camera obtained by PnP, stamped as $\tau_t^k$ and $\tau_t^{k\smallplus 1}$ by the clock of the target camera;
$\transformhat{c_t^{k\smallplus 1}}{c_t^{k}}$ is the relative pose of the target camera derived using the initialized continuous-time trajectory, and spatiotemporal parameters to be estimated.
At this stage, all parameters within the estimator have been rigorously initialized.

\subsection{Continuous-Time Batch Optimization}
\label{sect:batch_opt}
Finally, a continuous-time-based bundle adjustment would be performed to refine all initialized parameters to the optimal states.
The 3D-2D correspondences, organized from tracked grid patterns of the reference and target cameras, would be involved in constructing visual projection residuals for spatiotemporal optimization.
The corresponding nonlinear least-squares problem can be expressed as follows:
\begin{equation}
\small
\label{equ:ba}
\hat{\mathcal{X}}\gets\arg\min
\sum_{k}^{\mathcal{P}_{\mathrm{ref}}}\sum_{j}^{\mathcal{G}^k_{\mathrm{ref}}}
\rho\left(\left\|\bsm{e}_{\mathrm{ref}}^{k,j}\right\|^2\right)+
\sum_{k}^{\mathcal{P}_{\mathrm{tar}}}\sum_{j}^{\mathcal{G}^k_{\mathrm{tar}}}
\rho\left(\left\|\bsm{e}_{\mathrm{tar}}^{k,j}\right\|^2\right)
\end{equation}
with
\begin{equation}
\small
\begin{aligned}
\bsm{e}_{\mathrm{ref}}^{k,j} &=
\pi\left(\left( \transformhat{c_r}{w}( \tau^k_j) \right)^{-1}\cdot \bsm{p}^w_j,\mathcal{X}_{\mathrm{intr,ref}} \right) -\tilde{\bsm{x}}^k_j
\\
\bsm{e}_{\mathrm{tar}}^{k,j} &=
\pi\left(\left( \transformhat{c_r}{w}( \tau^k_j+\timeoffsethat{c_t}{c_r})\cdot\transformhat{c_t}{c_r} \right)^{-1}\cdot \bsm{p}^w_j,\mathcal{X}_{\mathrm{intr,tar}} \right) -\tilde{\bsm{x}}^k_j
\end{aligned}
\end{equation}
where $\bsm{e}_{\mathrm{ref}}^{k,j}$ and $\bsm{e}_{\mathrm{tar}}^{k,j}$ denote the projection residuals of 3D-2D pairs $\{\bsm{p}^w_j,\tilde{\bsm{x}}^k_j\}$ of reference and target cameras, respectively;
$\mathcal{X}_{\mathrm{intr,ref\mid tar}}$ are the intrinsics of two cameras;
$\transformhat{c_r}{w}(\tau)$ is the pose of $\coordframe{c_r}$ in $\coordframe{w}$ at time $\tau$, computed using the rotation and position B-splines;
$\pi(\cdot)$ represents the visual projection function, which has been defined in (\ref{equ:proj_func});
$\rho(\cdot)$ is the Huber loss function \cite{huber1992robust}.
The nonlinear least-squares problems, i.e., (\ref{equ:spline_init}), (\ref{equ:st_init}), and (\ref{equ:ba}), would be solved using \emph{Ceres} \cite{Agarwal_Ceres_Solver_2022}.

\section{Real-World Experiment}
To validate the feasibility of the proposed \emph{eKalibr-Stereo} and evaluate its performance, comprehensive real-world experiments were conducted.

\begin{figure}[t]
	\centering
	\includegraphics[width=\linewidth]{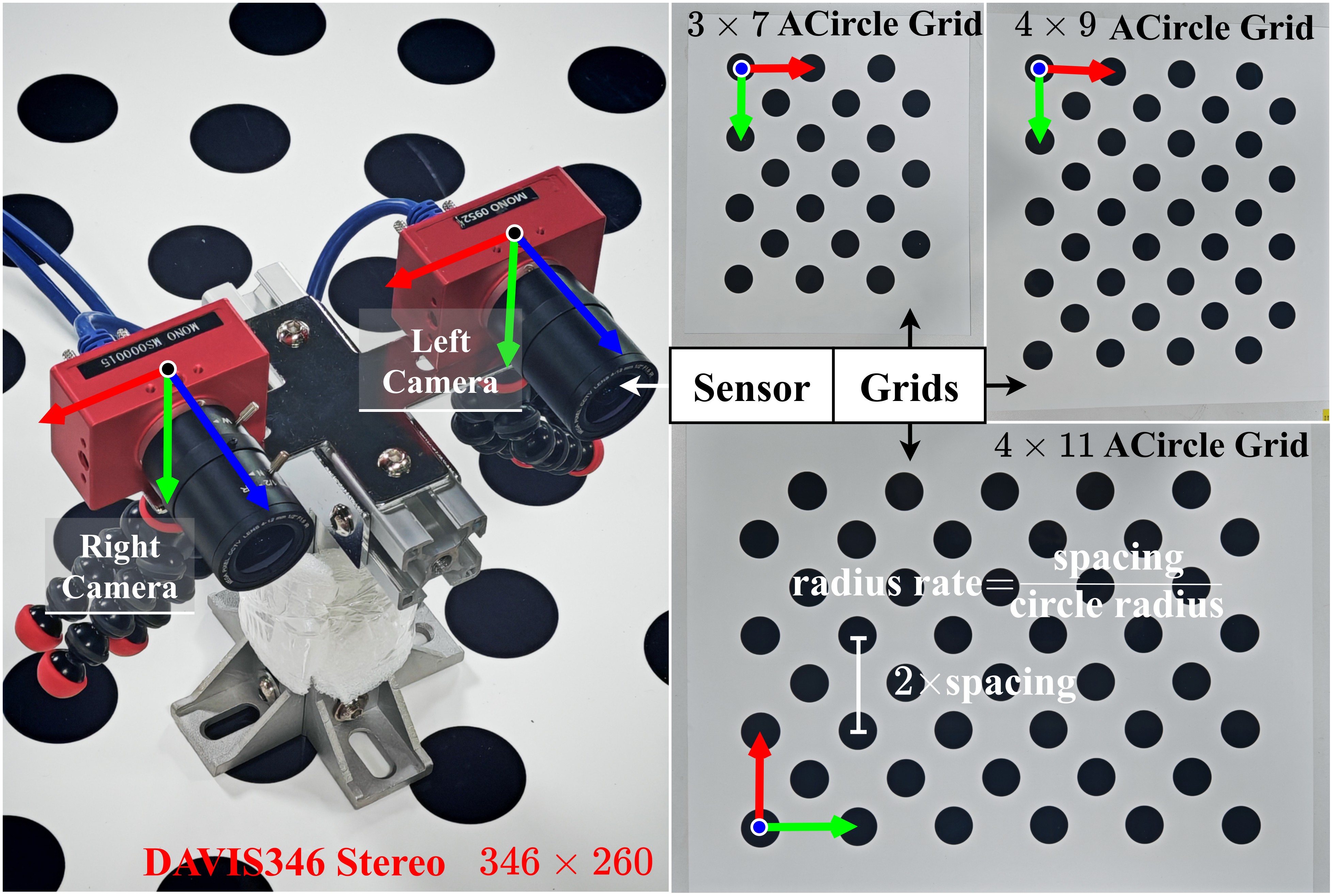}
	\caption{Stereo event camera rig (left subfigure) and three kinds of asymmetric circle grid patterns (right subfigures) utilized in real-world experiments.}
	\label{fig:setup}
\end{figure}

\subsection{Equipment Setup}
Fig. \ref{fig:setup} shows the self-assembled sensor suite for real-world experiments, consisting of two hardware-synchronized \emph{DAVIS346} event cameras (the resolution is 346$\times$260).
We refer to the two event cameras as the left camera and the right camera for convenience in subsequent description and discussion.
To ensure the comprehensiveness of the experiment, three different sizes of asymmetric circle grid patterns (3$\times$7, 4$\times$9, and 4$\times$11), as shown in Fig. \ref{fig:setup}, are used in real-world experiments.
The radius rate and spacing for all grid boards are 2.5 and 50 mm, respectively.

\subsection{Evaluation and Comparison of Grid Pattern Tracking}
\begin{figure}[t]
	\centering
	\includegraphics[width=\linewidth]{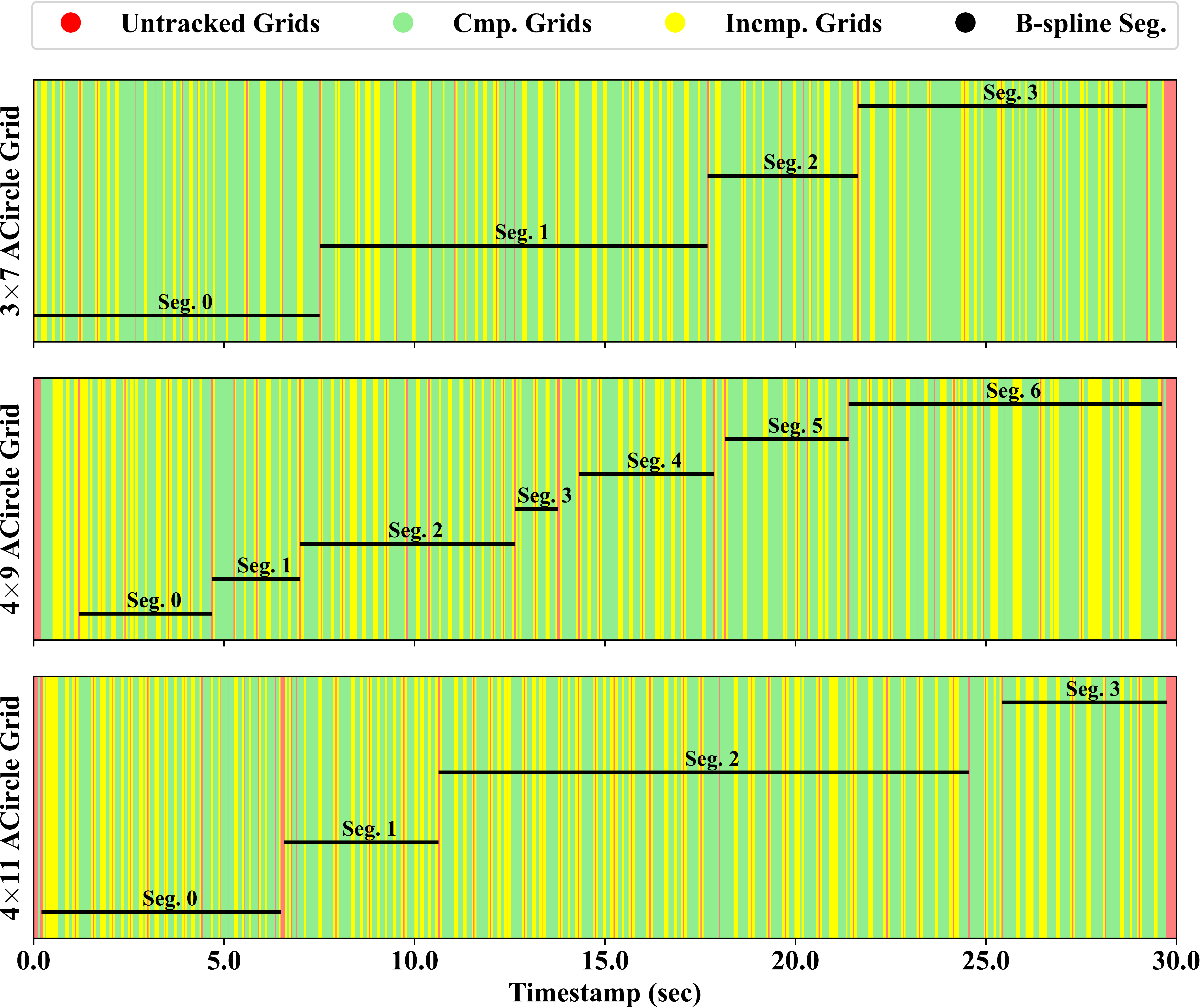}
	\caption{Grid tracking performance of \emph{eKalibr-Stereo} for three different sizes of ACircle grid boards.
	Based on the time distance threshold $\Delta\tau_{\mathrm{thd}}$ (set to 0.1 sec here) and the continuity threshold $\mathcal{N}_{\mathrm{thd}}$ (set to 50 here), tracked grids are segmented (black lines) and corresponding piece-wise B-splines would be constructed. Note that the total span of the segments is related to the tracking continuity, rather than the board size. More details are shown in Section \ref{sect:traj_init}.}
	\label{fig:grid_tracking}
	\figtabbottomspace
\end{figure}

\begin{table}[t]
\centering
\caption{\textbf{Evaluation and Comparison of Circle Grid Tracking}\\
eKalibr-Stereo achieves the highest tracking success rate}
\label{tab:grid_tracking}
\begin{threeparttable}
\begin{tabular}{c|l|ccc}
\toprule
Method                          & Grid Type         & Cmp. Grid & Incmp. Grid & Total    \\ \midrule
\multirow{3}{*}{eKalibr}        & \quad3$\times 7$  & 64.07 \%  & \ding{55}   & 64.07 \% \\
                                & \quad4$\times 9$  & 57.43 \%  & \ding{55}   & 57.43 \% \\
                                & \quad4$\times 11$ & 50.69 \%  & \ding{55}   & 50.69 \% \\ \midrule
\multirow{3}{*}{eKalibr-Stereo} & \quad3$\times 7$  & 64.07 \%  & 27.63 \%    & 91.69 \% \\
                                & \quad4$\times 9$  & 57.43 \%  & 28.48 \%    & 85.91 \% \\
                                & \quad4$\times 11$ & 50.69 \%  & 39.17 \%    & 89.86 \% \\ \bottomrule
\end{tabular}
\begin{tablenotes} 
\item[*] \emph{Cmp. Grid} and \emph{Incmp. Grid} refer to complete grids and incomplete grids, respectively. The tracking success rate is the ratio of successful grid trackings to the total grids.
\end{tablenotes}
\end{threeparttable}
\figtabbottomspace
\end{table}

The performance of the grid tracking module described in Section \ref{sect:grid_tracking} is first evaluated.
Fig. \ref{fig:grid_tracking} shows the tracking results of \emph{eKalibr-Stereo} in three runs using different sizes of grid boards, where both complete and incomplete grid patterns were plotted.
As can be seen, although \emph{eKalibr} \cite{chen2025ekalibr} is able to extract accurate complete grid patterns (those green ones), the corresponding continuity of tracking is poor.
Building upon \emph{eKalibr}, the proposed \emph{eKalibr-Stereo} leverages the prior knowledge of motion continuity, to predict and track incomplete patterns using Lagrange polynomial, significantly improving the continuity of grid tracking.
Based on these continuously tracked patterns, piece-wise B-splines can be effectively constructed (black lines), providing the necessary foundation for subsequent spatiotemporal optimization.

Table \ref{tab:grid_tracking} further quantitatively summarizes the average rate of grid pattern tracking of \emph{eKalibr} and \emph{eKalibr-Stereo} in Monte-Carlo experiments.
It can be seen that \emph{eKalibr} achieves a relatively low pattern tracking rate, with an average of 57 \%, which is primarily caused by high-dynamic motion required by spatiotemporal estimation (to ensure parameter observability).
In contrast, \emph{eKalibr-Stereo} can track incomplete grid patterns, significantly improving the tracking rate (with an average of 88 \%, an increase of approximately 30 \% compared with \emph{eKalibr}) and corresponding tracking continuity.

\subsection{Evaluation and Comparison of Calibration Performance}
To comprehensively and quantitatively evaluate the spatiotemporal calibration performance of the proposed \emph{eKalibr-Stereo}, we conducted real-world Monte-Carlo experiments, where 5 sequences of 30-second data are collected for each grid board for evaluation.
A total of four stereo visual calibration methods were incorporated in experiments for the evaluation and comparison of calibration results:
\begin{enumerate}
\item \textbf{Frame-Based Stereo Calibration} (\textbf{DV Software} \cite{dv_software}):
The stereo frame-based (standard) visual extrinsic calibration toolkit provided by \emph{iniVation}, i.e., the developer of \emph{DAVIS346} event camera employed in our real-world experiments.
Since the \emph{DAVIS346} event camera supports standard frame output, stereo visual extrinsics can be accurately determined using the conventional frame-oriented calibration pipeline in \emph{DV Software}.
Therefore, the calibration results can be treated as the reference.

\item \textbf{Event-Based Image Reconstruction} (\textbf{E2VID} \cite{rebecq2019high}) \& \textbf{Kalibr} \cite{furgale2013unified}:
To calibrate event cameras (such as \emph{DVXplorer} \cite{DVXplorerGuide}) that only support event output, a common approach is to ($i$) first use an event-based frame reconstruction algorithm to generate standard images, and then ($ii$) perform calibration using these reconstructed images in conventional calibrator.
In our experiments, the event-based frame reconstruction method \emph{E2VID} \cite{rebecq2019high} and the well-known frame-based visual calibrator \emph{Kalibr} \cite{furgale2013unified} are utilized.

\item \textbf{eKalibr-Stereo without Incomplete Grid Tracking}: The proposed event-based stereo visual calibrator, supports both spatial (extrinsics) and temporal (time offset) determination. Note that only the extracted complete grids are used for solving in this method.

\item \textbf{eKalibr-Stereo with Incomplete Grid Tracking}: The proposed event-based stereo spatiotemporal calibrator. Note that both the extracted complete grids and tracked incomplete grids are used for solving in this method.
\end{enumerate}
Meanwhile, to ensure the evaluation of temporal calibration of \emph{eKalibr-Stereo} in experiments, we manually \textbf{shifted} the timestamps of all events generated by the \textbf{right camera} by $\Delta\tau_{\mathrm{shift}}$ after data acquisition, to simulate a stereo visual system with a time offset, i.e., we have $\tau_{\mathrm{right}}\gets\tau_{\mathrm{right}}-\Delta\tau_{\mathrm{shift}}$.
Therefore, the time offset of the right camera with respect to the left camera, i.e., $\timeoffset{\mathrm{right}}{\mathrm{left}}$, is theoretically equal to $\Delta\tau_{\mathrm{shift}}$.
$\Delta\tau_{\mathrm{shift}}$ is sequentially set to various values in the experiment, namely: \textbf{10 ms}, \textbf{20 ms}, \textbf{50 ms}, and \textbf{100 ms}, for a comprehensive evaluation.
Note that \emph{DV Software} only supports spatial (extrinsic) calibration, requiring the stereo event camera rig to be temporally synchronized, and thus are not considered in time offset evaluation.

\begin{table*}[t]
\centering
\caption{\textbf{Spatiotemporal Calibration Results in Monte-Carlo Experiments}
\\eKalibr-Stereo achieves calibration accuracy and reliability comparable to conventional frame-based DV Software
}
\label{tab:st_calib_results}
\begin{threeparttable}
\begin{tabular}{c|rrrrrr|rr}
\toprule
                                                                                             & \multicolumn{6}{c|}{Extrinsic}                                                                                                                                                                                                                                                     & \multicolumn{2}{c}{Temporal}                                  \\ \cmidrule{2-9} 
                                                                                             & \multicolumn{3}{c|}{Rotation (Euler angles, unit: degree)}                                                                                        & \multicolumn{3}{c|}{Translation (unit: cm)}                                                                                    & \multicolumn{2}{c}{Time Offset (unit: ms)}                    \\ \cmidrule{2-9} 
\multirow{-3}{*}{Method}                                                                     & \multicolumn{1}{c}{Roll}                 & \multicolumn{1}{c}{Pitch}               & \multicolumn{1}{c|}{Yaw}                                     & \multicolumn{1}{c}{X}                    & \multicolumn{1}{c}{Y}                    & \multicolumn{1}{c|}{Z}                   & \multicolumn{1}{c}{Est.}      & \multicolumn{1}{c}{Ref.}      \\ \midrule
DV Software                                                                                  & \cellcolor[HTML]{DCDCDC}-0.447$\pm$0.013 & \cellcolor[HTML]{DCDCDC}0.690$\pm$0.017 & \multicolumn{1}{r|}{\cellcolor[HTML]{DCDCDC}0.383$\pm$0.017} & \cellcolor[HTML]{DCDCDC}12.073$\pm$0.017 & \cellcolor[HTML]{DCDCDC}-0.010$\pm$0.010 & \cellcolor[HTML]{DCDCDC}-0.390$\pm$0.047 & \multicolumn{1}{c}{\ding{55}} & 0.0                           \\ \midrule
                                                                                             & -0.719$\pm$0.165                         & 0.471$\pm$0.426                         & \multicolumn{1}{r|}{0.665$\pm$0.394}                         & 13.203$\pm$0.883                         & -1.172$\pm$1.130                         & 1.174$\pm$1.745                          & 7.036$\pm$2.379               & \cellcolor[HTML]{DCDCDC}10.0  \\
                                                                                             & -0.725$\pm$0.138                         & 0.480$\pm$0.369                         & \multicolumn{1}{r|}{0.655$\pm$0.453}                         & 13.378$\pm$0.827                         & -1.092$\pm$1.204                         & 1.156$\pm$1.778                          & 17.013$\pm$2.336              & \cellcolor[HTML]{DCDCDC}20.0  \\
                                                                                             & -0.724$\pm$0.150                         & 0.496$\pm$0.380                         & \multicolumn{1}{r|}{0.661$\pm$0.409}                         & 13.301$\pm$0.859                         & -1.140$\pm$1.181                         & 1.121$\pm$1.750                          & 47.040$\pm$2.380              & \cellcolor[HTML]{DCDCDC}50.0  \\
\multirow{-4}{*}{E2VID \& Kalibr}                                                            & -0.731$\pm$0.173                         & 0.494$\pm$0.352                         & \multicolumn{1}{r|}{0.663$\pm$0.426}                         & 13.306$\pm$0.790                         & -1.289$\pm$1.200                         & 1.141$\pm$1.764                          & 97.033$\pm$2.354              & \cellcolor[HTML]{DCDCDC}100.0 \\ \midrule
                                                                                             & -0.379$\pm$0.102                         & 0.603$\pm$0.097                         & \multicolumn{1}{r|}{0.352$\pm$0.092}                         & 11.980$\pm$0.084                         & 0.075$\pm$0.076                          & -0.359$\pm$0.098                         & 9.613$\pm$0.300               & \cellcolor[HTML]{DCDCDC}10.0  \\
                                                                                             & -0.372$\pm$0.108                         & 0.604$\pm$0.100                         & \multicolumn{1}{r|}{0.353$\pm$0.094}                         & 11.979$\pm$0.082                         & 0.076$\pm$0.079                          & -0.357$\pm$0.096                         & 19.629$\pm$0.302              & \cellcolor[HTML]{DCDCDC}20.0  \\
                                                                                             & -0.379$\pm$0.112                         & 0.600$\pm$0.099                         & \multicolumn{1}{r|}{0.359$\pm$0.095}                         & 11.984$\pm$0.083                         & 0.073$\pm$0.078                          & -0.358$\pm$0.099                         & 49.720$\pm$0.302              & \cellcolor[HTML]{DCDCDC}50.0  \\
\multirow{-4}{*}{\begin{tabular}[c]{@{}c@{}}eKalibr-Stereo\\ (w/o Incmp. Grid)\end{tabular}} & -0.371$\pm$0.103                         & 0.602$\pm$0.090                         & \multicolumn{1}{r|}{0.351$\pm$0.093}                         & 11.982$\pm$0.084                         & 0.073$\pm$0.075                          & -0.357$\pm$0.097                         & 99.822$\pm$0.301              & \cellcolor[HTML]{DCDCDC}100.0 \\ \midrule
                                                                                             & -0.430$\pm$0.053                         & 0.685$\pm$0.031                         & \multicolumn{1}{r|}{\textbf{0.392$\pm$0.004}}                & \underline{12.076$\pm$0.012}             & 0.003$\pm$0.016                          & -0.371$\pm$0.078                         & \textbf{9.984$\pm$0.195}      & \cellcolor[HTML]{DCDCDC}10.0  \\
                                                                                             & -0.430$\pm$0.053                         & 0.685$\pm$0.031                         & \multicolumn{1}{r|}{0.393$\pm$0.004}                         & 12.076$\pm$0.012                         & \underline{0.003$\pm$0.015}              & \textbf{-0.378$\pm$0.081}                & 19.962$\pm$0.195              & \cellcolor[HTML]{DCDCDC}20.0  \\
                                                                                             & \underline{-0.431$\pm$0.053}             & \textbf{0.686$\pm$0.030}                & \multicolumn{1}{r|}{\underline{0.392$\pm$0.004}}             & \textbf{12.076$\pm$0.011}                & \textbf{0.002$\pm$0.014}                 & -0.375$\pm$0.086                         & \underline{49.976$\pm$0.195}  & \cellcolor[HTML]{DCDCDC}50.0  \\
\multirow{-4}{*}{\begin{tabular}[c]{@{}c@{}}eKalibr-Stereo\\ (w/ Incmp. Grid)\end{tabular}}  & \textbf{-0.432$\pm$0.051}                & \underline{0.685$\pm$0.028}             & \multicolumn{1}{r|}{0.392$\pm$0.005}                         & 12.076$\pm$0.012                         & 0.003$\pm$0.015                          & \underline{-0.378$\pm$0.087}             & 99.928$\pm$0.196              & \cellcolor[HTML]{DCDCDC}100.0 \\ \bottomrule
\end{tabular}
\begin{tablenotes} 
\item[*] All spatiotemporal parameters in this table, i.e., extrinsics and time offset, are those of the right camera with respect to the left camera.
\item[*] The value in each table cell is represented as (Estimate Mean) $\pm$ (STD). A smaller STD indicates better repeatability and stability of the method.
\item[*] The best results are highlighted in \textbf{bold}, while the second-best results are \underline{underlined}, under the condition where the extrinsics from DV Software and the simulated time delays are used as references (i.e., values with \colorbox[HTML]{DCDCDC}{gray background}).
\end{tablenotes}
\end{threeparttable}
\figtabbottomspace
\end{table*}

Table \ref{tab:st_calib_results} summarized final spatiotemporal calibration results of \emph{DV Software}, \emph{E2VID} \& \emph{Kalibr}, and \emph{eKalibr-Stereo (w/o and w/ Incmp. Grid)} in real-world Monte-Carlo experiments, showing the spatiotemporal estimates and corresponding standard deviations (STDs).
As can be seen, the calibration method using \emph{E2VID \& Kalibr} achieved the poorest results with the largest STDs and biases.
Compared to the values from frame-based \emph{DV Software}, the estimated values from \emph{E2VID \& Kalibr} deviated by approximately 0.3 degrees for extrinsic rotation (with a STD of 0.3 degrees), 1.0 cm for extrinsic translation (with a STD of 1.2 cm), and 3.0 ms for time offset (with a STD of 2.3 ms).
This is mainly due to the high noise in reconstructed image frames from \emph{E2VID} (though reconstructed images are consistent on a macroscopic scale), which could reduce the extraction accuracy of the grid board in \emph{Kalibr}, further affecting the spatiotemporal determination.
In comparison, \emph{eKalibr-Stereo} is able to directly and accurately extract grid patterns from raw events, thereby achieving calibration results comparable to frame-based \emph{DV Software}.
Specifically, in the case of \emph{eKalibr-Stereo} with and without incomplete grid tracking, \emph{eKalibr-Stereo} with incomplete grid tracking is able to track more grid boards by utilizing motion priors, thereby providing stronger and more continuous motion constraints. As a result, it achieves better calibration results (closer to the estimates from \emph{DV Software}) and improved repeatability (with a smaller STD), compared to \emph{eKalibr-Stereo} without incomplete grid tracking.
Overall, among the three event-based methods, eKalibr-Stereo with incomplete grid tracking achieves the best results, with the closest match to DV Software's results and the smallest STD.

\begin{figure}[t]
	\centering
	\includegraphics[width=\linewidth]{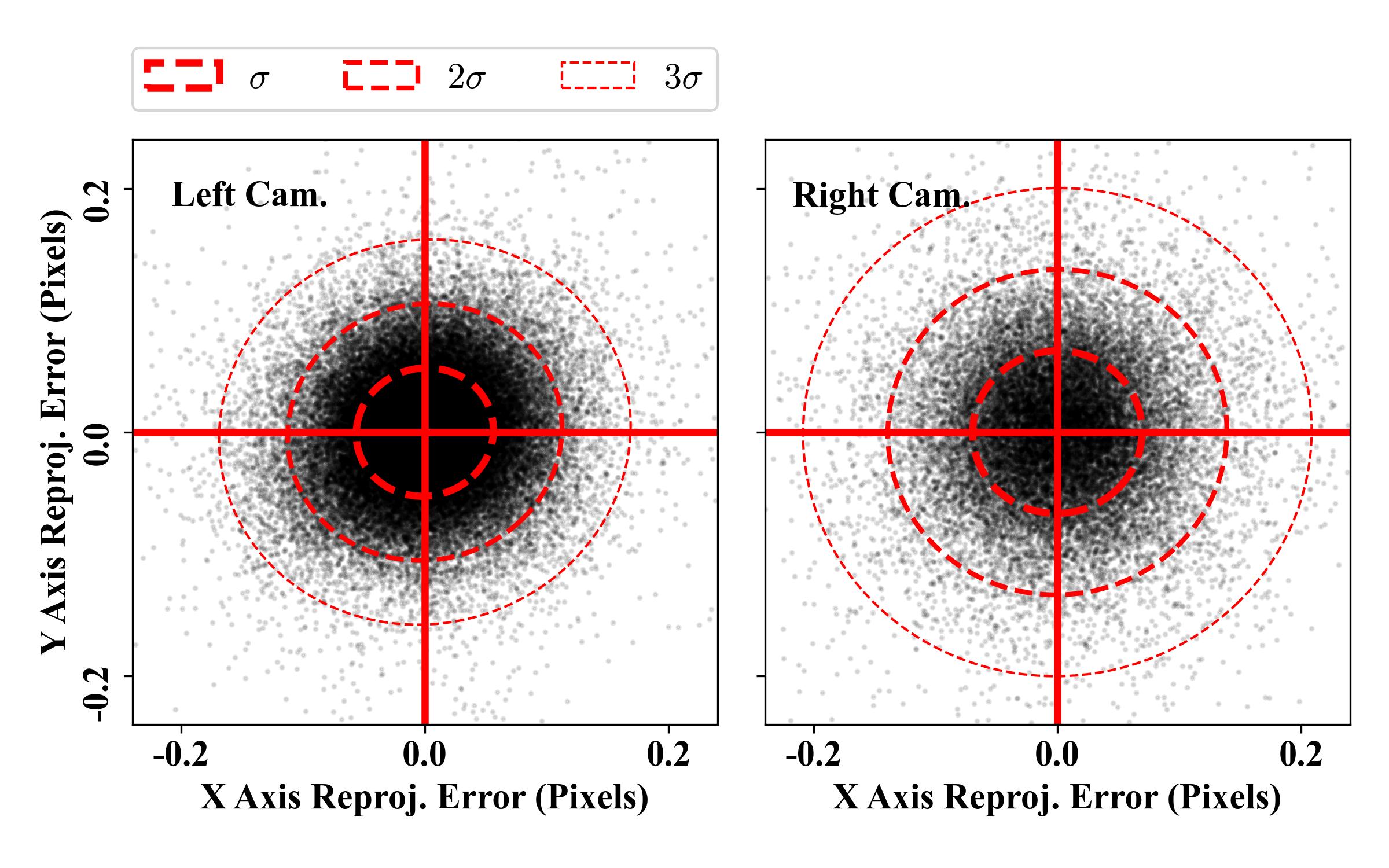}
	\caption{The distributions of projection errors after calibration.
	$\sigma$, $2\sigma$, and $3\sigma$ represent one, two, and three STDs of the reprojection errors, respectively.
	}
	\label{fig:proj_error}
	\figtabbottomspace
\end{figure}

Fig. \ref{fig:proj_error} plotted the distributions of projection errors of two cameras after spatiotemporal calibration in \emph{eKalibr-Stereo}.
It can be found that the projection errors follow a zero-mean normal distribution, indicating that the calibrated spatiotemporal parameters are well-estimated and unbiased.
The average sigma of final projection errors is less than 0.1 pixels, indicating the high-accuracy spatiotemporal calibration \emph{eKalibr-Stereo} capable of.

\subsection{Evaluation of Computation Consumption}

\begin{table}[t]
\centering
\caption{\textbf{Computation Consumption in eKalibr-Stereo}\\
Grid tracking consumed the majority of the processing time}
\tabtitlespace
\label{tab:computation}
\begin{threeparttable}
\begin{tabular}{c|cllcc}
\toprule
\multirow{3}{*}{Config.}               & OS Name               & \multicolumn{4}{l}{Ubuntu 20.04.6 LTS 64-Bit}                            \\ \cmidrule{2-6} 
                                       & Processor             & \multicolumn{4}{l}{12th Gen Intel® Core™ i9}                             \\ \cmidrule{2-6} 
                                       & Graphics              & \multicolumn{4}{l}{Mesa Intel® Graphics}                                 \\ \midrule\midrule
\multirow{2}{*}{Grid Type}             & \multicolumn{5}{c}{Computation Consumption (\textbf{unit: minute})}                              \\ \cmidrule{2-6} 
                                       & \multicolumn{3}{c|}{Grid Tracking (Cmp. + Incmp.)} & \multicolumn{1}{c|}{Optimization}  & Total  \\ \midrule
\multicolumn{1}{l|}{\quad3$\times 7$}  & \multicolumn{3}{c|}{2.47 + 2.29}                   & \multicolumn{1}{c|}{0.21}          & 5.00   \\
\multicolumn{1}{l|}{\quad4$\times 9$}  & \multicolumn{3}{c|}{2.65 + 2.64}                   & \multicolumn{1}{c|}{0.25}          & 5.54   \\
\multicolumn{1}{l|}{\quad4$\times 11$} & \multicolumn{3}{c|}{2.89 + 2.60}                   & \multicolumn{1}{c|}{0.31}          & 5.80   \\ \bottomrule
\end{tabular}
\begin{tablenotes} 
\item[*] The reported time represents the average time consumption across multiple (five) runs, each data sequence lasting 30 sec.
\item[*] The reported time in \textbf{Grid Tracking} means the average elapsed time for the left camera and right camera.
\end{tablenotes}
\end{threeparttable}
\figtabbottomspace
\end{table}

To evaluate the computation efficiency of \emph{eKalibr-Stereo}, we recorded the runtime for each execution in the Monte-Carlo experiments and calculated the average time consumption. The corresponding results are summarized in Table \ref{tab:computation}.
It can be observed that \emph{eKalibr-Stereo} takes approximately 5.5 minutes on average to calibrate a stereo event camera rig.
In the calibration process, the majority of the time, approximately 90\%, is spent on ACircle grid pattern extraction and tracking.
The total time consumption increases with the size of the grid, which is reasonable.

\section{Conclusion} 
In this article, we present the proposed continuous-time-based spatiotemporal calibrator for event-based stereo visual systems, named \emph{eKalibr-Stereo}, which is event-only and can accurately estimate both extrinsic and temporal parameters of the sensor suite.
To improve the continuity of grid tracking, building upon \emph{eKalibr}, an additional efficient procedure is designed in \emph{eKalibr-Stereo} to track incomplete grid patterns.
Based on tracked complete and incomplete grid patterns, a two-step initialization is first performed to recover the initial guesses of all parameters in the estimator, followed by a continuous-time batch optimization to refine all parameters to the optimal states.
Extensive real-world experiments were conducted to evaluate the performance of the \emph{eKalibr-Stereo} regarding grid tracking and spatiotemporal calibration.
The results indicate that \emph{eKalibr-Stereo} significantly improves the event-based grid tracking rate and could achieve spatiotemporal calibration accuracy comparable to frame-based stereo visual calibrators.


\section*{CRediT Authorship Contribution Statement}
\label{sect:author_contribution}
\textbf{Shuolong Chen}: Conceptualisation, Methodology, Software, Validation, Original Draft.
\textbf{Xingxing Li}: Supervision, Funding Acquisition, Review and Editing.
\textbf{Liu Yuan}: Data Curation, Review and Editing.
	
\bibliographystyle{IEEEtran}
\bibliography{reference}

\end{document}